\documentclass[preprint,12pt]{elsarticle}

\usepackage[T1]{fontenc}
\usepackage{hyperref}

\usepackage{amssymb}
\usepackage{algorithm}
\usepackage{algpseudocode}
\usepackage{rotating}
\usepackage{xcolor}

\journal{Engineering Applications of Artificial Intelligence}

\begin{document}

\begin{frontmatter}



\title{Explained anomaly detection in text reviews: Can subjective scenarios be correctly evaluated?}


\author[inst1]{David Novoa-Paradela\corref{cor1}}
\ead{david.novoa@udc.es}

\affiliation[inst1]{organization={Universidade da Coruña, CITIC},
            addressline={Campus de Elviña s/n}, 
            postcode={15008}, 
            city={A Coruña},
            country={Spain}}

\author[inst1]{Oscar Fontenla-Romero}
\ead{oscar.fontenla@udc.es}

\author[inst1]{Bertha Guijarro-Berdiñas}
\ead{berta.guijarro@udc.es}

\cortext[cor1]{Corresponding author}

\begin{abstract}

This paper presents a pipeline to detect and explain anomalous reviews in online platforms. The pipeline is made up of three modules and allows the detection of reviews that do not generate value for users due to either worthless or malicious composition. The classifications are accompanied by a normality score and an explanation that justifies the decision made. The pipeline's ability to solve the anomaly detection task was evaluated using different datasets created from a large Amazon database. Additionally, a study comparing three explainability techniques involving 241 participants was conducted to assess the explainability module. The study aimed to measure the impact of explanations on the respondents' ability to reproduce the classification model and their perceived usefulness. This work can be useful to automate tasks in review online platforms, such as those for electronic commerce,  and offers inspiration for addressing similar problems in the field of anomaly detection in textual data.  We also consider it interesting to have carried out a human evaluation of the capacity of different explainability techniques in a real and infrequent scenario such as the detection of anomalous reviews, as well as to reflect on whether it is possible to explain tasks as humanly subjective as this one.

\end{abstract}




\end{frontmatter}

\color{black}
\section{Introduction}
\label{sec:introduction}

Nowadays more than ever in history, the user opinions about products and services have a great impact on the future of the company that offers them. In such a globalized and highly competitive world, online review platforms, such as electronic commerce (e-commerce), play a crucial role in the credibility of products and services. These user reviews usually come in the form of text reviews or numerical ratings, accompanied in some cases by images or videos. These reviews provide users with information about the product or service they are considering purchasing, which directly influences the number of sales. Most people make purchase decisions based on ratings and reviews from other users \cite{VONHELVERSEN20181}.

In the case of many companies, such as Amazon, each product in the store has a list of text reviews published by customers of the platform. Users can access this list of reviews (opinions) to obtain extra information about the product, being able to mark the reviews as useful, which will position those reviews with the most votes at the top of the list. In addition to this, users can report to Amazon if they feel a review is inappropriate, for example, if its content is incorrect. This procedure to rank reviews based on their usefulness and report inappropriate reviews is carried out manually by platform users. As a result, Amazon reported more than 200 million suspected fake reviews in 2020 alone \cite{amazon}. This problem does not only occur on Amazon, but affects all platforms that allow their users to post reviews. For example, Tripadvisor uses an automatic system capable of distinguishing between normal, suspicious and inappropriate reviews \cite{tripadvisor}. Inappropriate ones are automatically removed (3.1\% of review submissions in 2020), while those classified as suspicious are reviewed again by a human moderator (5.1\% of review submissions in 2020).

On the other hand, in machine learning (ML), anomaly detection (AD) is the branch that builds models capable of differentiating between normal and abnormal data \cite{chandola}. At first, anomaly detection might seem like a classification problem with only two classes. However, anomalies tend to occur infrequently or are non-existent, so normal data prevails in these scenarios. Because of this, it is common for models to be trained only with normal data. The goal of these models is to represent the normal class as well as possible in order to classify new data as normal or anomalous.

The technological development of recent years has allowed the construction of very powerful models for Natural Language Processing (NLP) \cite{DBLP:journals/corr/abs-2105-00813}. Contrary to other tasks such as Sentiment Analysis \cite{TABINDAKOKAB2022100157} or Question Answering \cite{KIM2022104061}, the application of anomaly detection on texts is still at an early stage, probably due to its lower demand. For this reason, we present in this article a pipeline that, given text reviews of a product (in this case from Amazon), addresses opinion filtering as an anomaly detection problem where:

\begin{itemize}
    \item Reviews containing representative information about the product are considered as the normal class.
    \item Reviews whose content has nothing to do with the product to be represented are considered as the anomalous class.
\end{itemize}

The proposed pipeline allows us to carry out the following tasks:

\begin{itemize}
    \item Classify reviews as normal or anomalous, allowing us to locate those that do not describe characteristics of the product to which they are associated, and therefore have no value for the users of the platform.
    \item Issue a normality score associated with each review.
    \item Generate an explanation that justifies the classification made for each review by the system.
\end{itemize}

The pipeline's ability to solve the anomaly detection task was evaluated using different datasets created from a large Amazon database \cite{amazon_data}. In addition, to evaluate the explainability module, a study was carried out to compare three explainability techniques in which a total of 241 people had participated. The objective of this study is both to measure the impact of the explanations on the reproducibility of the classification model by the respondents, as well as the usefulness of these explanations. 

We believe that this work can be useful to automate tasks such as those mentioned in online review platforms, in addition to the existing interest in the application of anomaly detection models on texts, for which it can serve as inspiration to solve similar problems. We also consider it interesting to have carried out a human evaluation of the capacity of different explainability techniques in a real and infrequent scenario such as the detection of anomalous reviews, as well as to reflect on whether it is possible to explain tasks as humanly subjective as this one.

This document is structured as follows. Section 2 contains a brief review of the main anomaly detection works in texts and, more  specifically, in text reviews. Section 3 describes the proposed pipeline and its operation. Section 4 collects the experimentation carried out and, finally, conclusions are drawn in Section 5.

\section{Related work}\label{relatedwork}


Anomaly detection has been a consolidated research field for years. Its great utility has allowed its techniques to be applied in numerous areas: medicine \cite{SCHNEIDER2022193}, industrial systems \cite{TRUONG2022103692}, electronic fraud \cite{HILAL2022116429}, cybersecurity \cite{HUONG2021103509}, etc. However, when we talk about texts and NLP, there is no massive application of these anomaly detection techniques as in the previous cases. This may be due to the difficulty in defining the idea of an anomaly in texts. Contrary to other scenarios, such as monitoring an industrial system through its sensors, in which the anomalous class will correspond to faults in the system, when the data is text, defining the concept of an anomaly is not trivial.

One of the most important lines of research related to the detection of anomalies in texts is fake reviews detection, also known as spam review detection, fake opinion detection, and spam opinion detection \cite{9416474}. The main problem associated with fake review detection is classifying the review as either fake or genuine. There are generally three types of fake reviews \cite{10.1145/1341531.1341560}: 


\begin{itemize}
    \item \textbf{Type 1 (untruthful opinions)}: Fake reviews describing users who post negative reviews to damage a product's reputation or post positive reviews to promote it. These reviews are called fake or deceptive reviews, and they are difficult to detect simply by reading, as real and fake reviews are similar to each other.
    \item \textbf{Type 2 (reviews on brands only)}: Those that do not comment on the products themselves, but talk about the brands, manufacturers, or sellers of the products. Although they can be useful, they are sometimes considered spam because they are not targeted at specific products.
    \item \textbf{Type 3 (non-reviews)}: Non-reviews that are irrelevant and offer no genuine opinion.
\end{itemize}

In the work carried out by J. Salminen \textit{et al}. \cite{SALMINEN2022102771}, the authors try to distinguish genuine reviews from fake reviews on Amazon. To have a labeled dataset of fake reviews, they use GPT-2 \cite{radford2019language} to artificially generate them. After this, they solve the task of distinguishing between genuine and fake reviews by fine-tuning a pretrained RoBERTa \cite{DBLP:journals/corr/abs-1907-11692} model. They also show that his model is also capable of detecting fake reviews manually written by humans. Ş. Öztürk Birim \textit{et al}. \cite{BIRIM2022884} proposed to use relevant information as the review length, purchase verification, sentiment score, or topic distribution as features to represent costumers reviews. Based on these features, well-known machine learning classifiers like random forests (RF) \cite{breiman2001random} are applied for fake detection. In their article, D. U. Vidanagama \textit{et al}. \cite{VIDANAGAMA2022117869} incorporate review-related features such as linguistic features, Part-of-Speech (POS) features, and sentiment analysis features using a domain ontology to detect fake reviews with a rule-based classifier.

L. Ruff \textit{et al}. \cite{ruff-etal-2019-self} presented Context Vector Data Description (CVDD), a text-specific anomaly detection method that allows working with sequences of variable-length embeddings using self-attention mechanisms. To overcome the limitations of CVDD, J. Mu \textit{et al}. \cite{MU2021286} proposed tadnet, a textual anomaly detection network that uses an adversarial training strategy to detect anomalous texts in Social Internet of Things. In addition, thanks to the capture of the different semantic contexts of the texts, both models achieve interpretability and flexibility, allowing to detect which parts of the texts have caused the anomaly.

Other authors, instead of developing text-specific AD models, make use of well-known AD and NLP techniques to design architectures that solve the problem. B. Song \textit{et al}. \cite{SONG201947} propose to analyze the accident reports of a chemical processing plant to detect anomalous conditions. In this scenario, anomalous conditions are defined as unexperienced accidents that occur in unusual conditions. The authors work directly with the original text extracting the meaningful keywords of the reports using the term frequency-inverse document frequency (TF-IDF) index. Based on this, and using the local outlier factor (LOF) algorithm, they identify anomaly accidents in terms of local density clusters, finding four major types of anomaly accidents. Working with the original texts and not with embeddings they achieve a certain interpretability in the results. In the approach presented by S. Seo \textit{et al}. \cite{SEO2020113111}, a framework for identifying unusual but noteworthy customer responses and extracting significant words and phrases is proposed. The authors use Doc2Vec to vectorize customer responses, then LOF is applied to identify unusual responses, and based on a TF-IDF analysis and the distances in the embedding space, they can visualize useful information about the results through a network graph.

In the previous works and most cases, the detection of fake reviews focuses on detecting type 1 reviews, that is, reviews that positively or negatively describe a product but whose intention is not genuine since they do not come from a real buyer. Since there are several works based on this scenario \cite{SALMINEN2022102771, BIRIM2022884, VIDANAGAMA2022117869}, in this work we will focus on detecting type 2 and 3 reviews, which refer to reviews that do not provide information about the product itself. In this way, the objective of this work is to design a pipeline capable of distinguishing reviews related to a specific product (normal reviews) from reviews that do not and therefore do not provide information to the users that read them (anomalous reviews), for example, because they wrongly describe other products or because they are too generic. In addition to this, the classifications carried out by the system will be explained through an analysis process based on NLP techniques.

\section{The proposed pipeline} \label{method}

\begin{figure}
\centerline{\includegraphics[width=\textwidth]{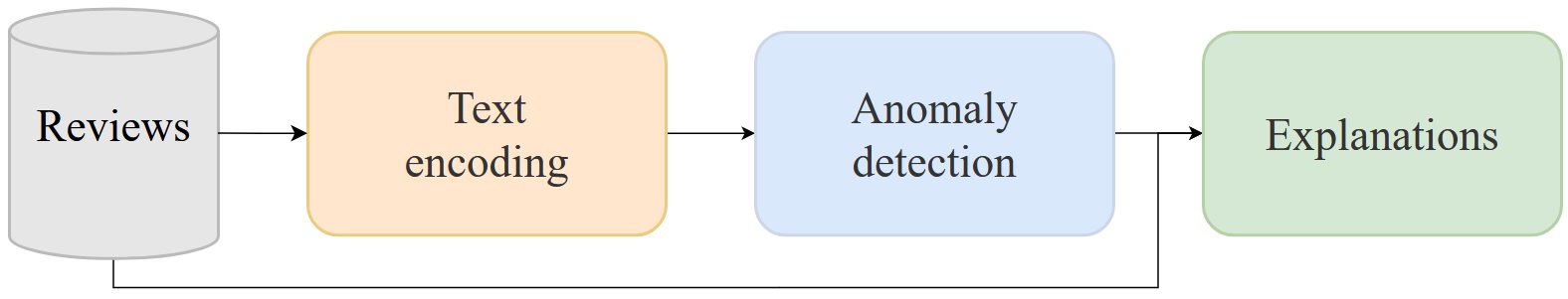}}
\caption{General modules that form the proposed pipeline.}
\label{figureAD_modulos_genericos}
\end{figure}

The purpose of the proposed pipeline is, given the text reviews of an Amazon product, to classify them as normal if they refer to it, and as anomalous if they describe different products or if they are so generic that they do not provide  useful information to consumers. These classifications will be accompanied by a normality score and explanations that justify them. Figure \ref{figureAD_modulos_genericos} shows the modules that are part of the proposed pipeline to solve this task. In the first phase, we encode the text reviews of the target product using the pretrained MPNet transformer \cite{https://doi.org/10.48550/arxiv.2004.09297}. In the second phase, a DAEF \cite{NOVOAPARADELA2023109805} autoencoder is trained using these embeddings. Using the reconstruction errors issued by the network and a predefined threshold error, we can classify reviews as normal (error greater than the threshold) or anomalous (error less than the threshold). For the last phase, we propose a method based on the most frequent normal terms to generate an explanation associated with each classification.


Therefore, the proposed pipeline is composed of three main modules: (1) Text encoding; (2) Anomaly detection; (3) Explainability. The following sections describe each of them in detail.

\subsection{Text encoding} \label{method1}

Although there are models capable of dealing directly with images or text, machine learning models are usually designed to be trained using numerical vectors that represent the sample data (tabular data). This input format of the data is the usual one for classic anomaly detection methods. In the NLP area, there are multiple techniques to represent texts using vectors of real numbers, which are known as embeddings. These techniques allow the generation of vector spaces that try to represent the relationships and semantic similarities of the language, so that, for example, two synonymous words will be found at a shorter distance in the vector space than two unrelated words.

Embeddings can be calculated independently for each word of the language (word embeddings), which led to models such as word2vec \cite{https://doi.org/10.48550/arxiv.1301.3781} or Global Vectors (GloVe) \cite{pennington-etal-2014-glove}. The representation of a sentence (sentence embedding) or a document (document embedding) will therefore be the sum of all the individual representations of the terms that make it up. To obtain representations of fixed length, it is usual to perform operations such as the mean. In certain cases, these operations between embeddings can worsen or even invalidate the final embedding, so specific models have been developed capable of understanding and representing a text as a whole, instead of just encoding it word by word. Among these models are those based on transformers \cite{https://doi.org/10.48550/arxiv.1706.03762}, such as BERT \cite{https://doi.org/10.48550/arxiv.1810.04805}, XLNet \cite{https://doi.org/10.48550/arxiv.1906.08237}, GPT-3  \cite{https://doi.org/10.48550/arxiv.2005.14165} or GPT-4 \cite{openai2023gpt4}, which have been trained over large-scale datasets and can solve different tasks, including sentence embedding. 

Since it inherits the advantages of the BERT and XLNet models while overcoming their limitations, we will use the pre-trained MPNet \cite{https://doi.org/10.48550/arxiv.2004.09297} model to calculate the embeddings of the reviews. MPNet combines Masked Language Modeling (MLM) and Permuted Language Modeling (PLM) to predict token dependencies, using auxiliary position information as input to enable the model to view a complete sentence and reduce position differences. Models like GPT-3 or GPT-4 are more advanced, but in addition to not being open source, they demand much higher computational resources, unaffordable for a significant part of the scientific community, as is our particular case. 

MPNet maps sentences and paragraphs to a 768 dimensional dense vector space $\textbf{e} \in \mathbb{R}^{768\times 1}$, providing fast and quality encodings. Computing time is a critical aspect in the area in which this work is framed, since in e-commerce platforms (and online reviews in general) we can find a huge number of products and reviews to deal with. Therefore, the proposed pipeline uses MPNet to calculate the embeddings of the reviews, which allows them to be represented as numeric vectors of fixed size. This model and others are available in the Hugging Face repository \cite{hugging}.

\subsection{Anomaly detection} \label{method2}

Anomaly detection is a field with a large number of algorithms that solve the problem of distinguishing between normal and anomalous instances in a wide variety of ways \cite{chandola, khan}. Depending on the assumptions made and the processes they employ, we can distinguish between five main types of methods: probabilistic, distance-based, information theory-based, boundary-based, and reconstruction-based methods. Among the latter are autoencoder networks (AE) \cite{10.5555/1756006.1953039}, one of the most widely used models. AE are a type of self-associative neural network whose output layer seeks to reproduce the data presented to the input layer after having gone through a dimensional compression phase. In this way, they manage to obtain a representation of the input data in a space with a dimension smaller than the original, learning a compact representation of the data, retaining the important information, and compressing the redundant one. 

One of the requirements of our proposed pipeline is to generate a normality score for each review. This score will allow us, among other things, to order the different reviews by their level of normality. When AE networks are used in anomaly detection scenarios, the classification is usually carried out based on the reconstruction error that they emit to reproduce in its output the embeddings of the reviews it receives as inputs, which represents the level of normality of the evaluated instance. This reconstruction error can be used as the normality score we are looking for.

Due to the speed of its training, we have decided to use DAEF (Deep Autoencoder for Federated learning) \cite{NOVOAPARADELA2023109805} as our anomaly detection model. Unlike traditional neural networks, DAEF trains a deep AE network in a non-iterative way, which drastically reduces its training time without losing the performance of traditional (iterative) AEs. The proposed pipeline uses DAEF to calculate the embeddings of the reviews, being able to issue a normality score associated with each of the review classifications.

\subsection{Explainability} \label{method3}

As black-box ML models are mainly being employed to make important predictions in critical contexts, the demand for transparency is increasing. The danger is in creating and using decisions that are not justifiable, legitimate, or that do not allow obtaining detailed explanations of their behaviour. Explanations supporting the output of a model are crucial, moreover in fields such as medicine, autonomous vehicles, security, or finance.

When ML models do not meet any of the criteria imposed to declare them explainable, a separate method must be applied to the model to explain its decisions. This is the purpose of post-hoc explainability techniques \cite{BARREDOARRIETA202082}, which aim at communicating understandable information about how an already developed model produces its predictions for any given input. Within these techniques, the so-called model-agnostic techniques are those designed to be plugged into any model with the intent of extracting some information from its prediction procedure.

To generate the explanations we have implemented an approach based on a statistical analysis of the dataset. Based on the definitions of normal and anomalous review presented in Section \ref{relatedwork}, Our hypothesis assumes that normal reviews will always refer directly or indirectly to the target product so that there will be a list of terms used very frequently among normal texts. The appearance of one or more of these ``normal'' terms in a review would justify its classification by the system as normal. In the same way, anomalous reviews may be justified with the non-presence of said terms. 

\begin{figure}
\centerline{\includegraphics[width=35.7pc]{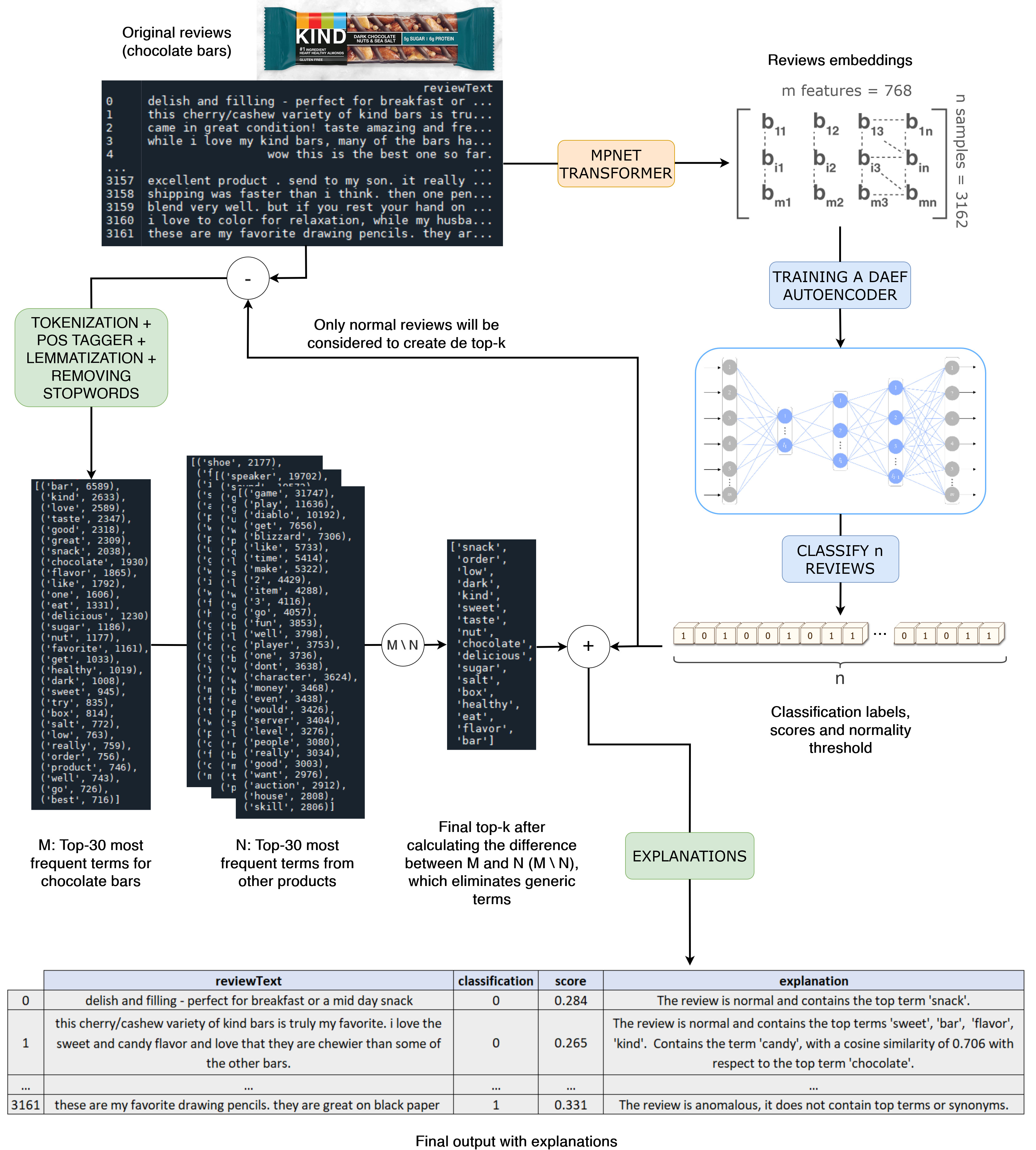}}
\caption{Proposed pipeline considering the product ``chocolate bars'' as the normal class.}
\label{arquitecturaCompleta}
\end{figure}

The original text reviews classified as normal using the AD model will be processed (tokenization, lemmatization and stopword removal) and analyzed to obtain the list of the top-n most frequent terms. This list of terms will be compared with the lists of frequent terms of other products in order to remove the terms they have in common, so that the final list for a product does not contain generic terms. This final list of terms will be used to generate the explanations. New reviews classified as normal will be searched for the presence of any of these terms. To make this search more flexible, it will also be analyzed if there are terms in the review semantically close to any of the words on the list. To quantify the closeness of two words, the cosine similarity emitted by the MPNet transformer model itself will be used as a measure of distance. Reviews classified as anomalous will be explained by the non-occurrence of terms from the list. Figure \ref{arquitecturaCompleta} represents an overall view of the pipeline considering the product ``chocolate bars'' as the normal class.

In summary, in this work, we propose the use of this statistical analysis of the terms to generate the explanations associated with the classification of the reviews.

\section{Evaluation} \label{results}

In this section several experiments are presented to show the behavior of the proposed pipeline in real scenarios. The section is divided into two parts: the evaluation of the anomaly detection task (Section \ref{evaluation1}), and the evaluation of explanations (Section \ref{evaluation2}). In the first one, the capability of the pipeline to detect anomalous reviews will be evaluated using products from the Amazon platform.  The second part will discuss the problems of evaluating explanations and will present a study involving humans to assess the benefits of adopting such explanations.

\subsection{Evaluating the anomaly detection task} \label{evaluation1}

The anomaly detection task that solves the proposed pipeline can be evaluated following the usual methodology in the field of anomaly detection. In this first part of the evaluation we will describe the test scenario used, the methodology employed, and we will discuss the results obtained.

\subsubsection{Experimental setup} \label{results1}

The objective of this study is to evaluate the capacity of the pipeline in a real anomaly detection scenario. Although Section \ref{method2} defines the requirements for the anomaly detection methods that the pipeline can use, in these tests different alternatives have been compared. Specifically, two non-iterative implementations of autoencoder networks and a boundary-based method have been employed. These methods are DAEF \cite{NOVOAPARADELA2023109805}, OS-ELM \cite{4012031}, and OC-SVM \cite{1437839} respectively.

For the evaluation, we employed datasets obtained from the large Amazon database \cite{amazon_data}, which collects reviews of various products from the year 1996 to 2018. The main problem with this dataset is that the reviews that compose it are not labeled as normal or anomalous. Because of this, in order to simulate a scenario similar to the one described throughout the article, we have decided to select the reviews corresponding to several of the most demanded products. Thus, for a given test we could consider the reviews of one product as normal, and introduce reviews from other products as anomalous. Five different product categories were selected, for which the two products with the highest number of reviews were used, resulting in a total of 10 products. Table \ref{tab:dataset} summarizes its characteristics. In all cases, MPNet was used as the model to encode the text reviews. 

Besides we considered two types of tests based on the products used:

\begin{itemize}
    \item \textbf{1vs.4 - Far products}: For each of the five categories, in this type of tests the product with the most reviews from one of the categories will be considered as the normal class, while the product with the highest number of reviews from each of the other four categories will be considered as the anomalous class. The fact that the product considered normal belongs to a different category should facilitate its distinction.
    \item \textbf{1vs.1 - Near products}: For each of the five categories, in these tests the two products with the most reviews of the same category will be selected. One will be considered as the normal class and the other as the abnormal class. The fact that both products belong to the same category should make it more difficult to distinguish them since they may have common characteristics. 
\end{itemize}

\begin{table}
\begin{center}
{\caption{Characteristics of the products used.}\label{tab:dataset}}
\resizebox{3.5in}{!}{%
\begin{tabular}{llr}
\\[-6pt]
\hline 
Product & Category & Reviews\\
\hline 
\href{https://www.amazon.com/dp/B00BUKL666}{Chocolate bars} & Grocery and Gourmet Food & 11526 \\
\href{https://www.amazon.com/dp/B00542YXFW}{Anise seeds} & Grocery and Gourmet Food & 9083 \\

\href{https://www.amazon.com/dp/B00006IEEV}{Colored pencils} & Office Products & 14340 \\
\href{https://www.amazon.com/dp/B00AE9V3WQ}{Ergonomic cushion} & Office Products & 11942 \\

\href{https://www.amazon.com/dp/B00HTK1NCS}{Gaming mouse} & Video Games & 6462 \\
\href{https://www.amazon.com/dp/B004RMK57U}{PS4 membership } & Video Games & 5135 \\

\href{https://www.amazon.com/dp/B010OYASRG}{Bluetooth speaker} & Electronics & 28539 \\
\href{https://www.amazon.com/dp/B00L0YLRUW}{Wi-Fi range extender} & Electronics & 20873 \\

\href{https://www.amazon.com/dp/B000V0IBDM}{Foot insoles} & Amazon Fashion & 4384 \\
\href{https://www.amazon.com/dp/B00I0VHS10}{Yoga leggings} & Amazon Fashion & 3889 \\
\hline
\end{tabular}
}
\end{center}
\end{table}

The anomaly detection algorithms were trained using only normal data (the product considered as normal), while the test phase included data from both classes in a balanced manner (50\% normal and 50\% anomalies). 

To evaluate the performance of each algorithm with each combination of hyperparameters a 10-fold has been used. The normal data were divided into 10-folds so that, at each training run, 9 folds of normal data were used for training, while the remaining fold and the anomaly set were used together in a balanced manner for testing. 


In this work, reconstruction errors at the output of the anomaly detector were calculated using Mean Squared Error (MSE). To establish the threshold above which this error indicates a given instance corresponds to an anomaly, among the various methods available,  we employed a popular approach based on the interquartile range (IQR) of the reconstruction errors of the training examples, defined by:


\begin{equation}
I\!Q\!R = Q_3 - Q_1 
\label{iqr}
\end{equation}

\noindent where $Q_1$ and $Q_3$ represent the first and the third quartiles. We define two error thresholds, one for outliers ($outlier I\!Q\!R$) and another for extreme outliers ($extreme I\!Q\!R$), as:

\begin{equation}
outlier I\!Q\!R = Q_3 + 1.5 \times I\!Q\!R
\label{iqr1}
\end{equation}

\begin{equation}
extreme I\!Q\!R = Q_3 + 3 \times I\!Q\!R 
\label{iqr2}
\end{equation}

In addition, throughout the tests, we also considered other thresholds using fixed percentiles ($Q_{95}$, $Q_{90}$, $Q_{80}$, $Q_{70}$, $Q_{60}$ and $Q_{50}$), since a priori it is not easy to figure out which one can provide the best results, so it is considered as an additional hyperparameter to be taken into account. The OC-SVM method constitutes an exception as it automatically assigns a score to each input instance and decides its classification based on an internally calculated threshold error.

To measure the performance of the algorithms the F1-score metric was used, considering the anomalous class as the positive one, and based on the number of true positives (TP), true negatives (TN), false positives (FP) and false negatives (FN):

\begin{equation}
F_1 = 2\frac{precision \cdot recall}{precision + recall} = \frac{2 \times TP}{2 \times TP+FP+FN}
\label{f1score}
\end{equation}


Finally, the combinations of hyperparameters chosen for each algorithm, as well as the error thresholds, were selected using a grid search and are available in Appendix A (Tables \ref{ta:dinos2} and \ref{ta:dinos3}).

All the evaluation tests were performed in a machine equipped with an Intel Core i7-11700k processor and 64GB of RAM.

\subsubsection{Evaluating the anomaly detection task} \label{results2}

Table \ref{ta:resultados1}  shows the results of the 1vs.4 - Far products test. As can be seen, the performance of the three models is remarkable, highlighting that obtained by the OS-ELM autoencoder network. We can affirm that the embeddings produced by the MPNet model provide an encoding with sufficient quality for the anomaly detection models to be able to differentiate between the evaluated products.
\begin{table}[H]
\centering
{\caption{Average test F1-score $\pm$ standard deviation for the 1vs.4 datasets.}\label{ta:resultados1}}
\vspace*{2mm}
\resizebox{5.4in}{!}{
\centering
\begin{tabular}{llccc}
\hline
Normal class &  \begin{tabular}{l} Anomalous class \end{tabular} & DAEF & OS-ELM & OC-SVM \\ 
\hline
Chocolate bars    & \begin{tabular}{l} Colored pencils, Gaming mouse, \\ Bluetooth speaker, Foot insoles \end{tabular} & 92.1$\pm$1.0 & 96.6$\pm$0.4  & 95.3$\pm$1.5  \\
Colored pencils    & \begin{tabular}{l} Chocolate bars, Gaming mouse, \\ Bluetooth speaker, Foot insoles \end{tabular} & 96.1$\pm$0.4 & 96.0$\pm$0.3  & 94.4$\pm$1.1  \\
Gaming mouse    & \begin{tabular}{l} Chocolate bars, Colored pencils, \\ Bluetooth speaker, Foot insoles \end{tabular} & 94.5$\pm$1.1 & 95.7$\pm$0.1  & 93.8$\pm$0.8  \\
Bluetooth speaker    & \begin{tabular}{l} Chocolate bars, Colored pencils, \\ Gaming mouse,  Foot insoles \end{tabular} & 95.7$\pm$0.8 & 96.9$\pm$0.2  & 94.6$\pm$0.3  \\
Foot insoles    & \begin{tabular}{l} Chocolate bars, Colored pencils, \\ Gaming mouse,  Bluetooth speaker \end{tabular} & 96.4$\pm$1.3 & 96.3$\pm$0.2  & 94.5$\pm$0.6  \\
\hline
\end{tabular}
}
\end{table}

Table \ref{ta:resultados2} collects the results of test 1vs.1 - Near products. As can be seen, the overall performance is still good, although it has been slightly reduced concerning the previous test, possibly because the task is a little more complicated as the products are closer together. Nevertheless, once again, the quality of the embeddings allows a proper differentiation between products.

\begin{table}[H]
\centering
{\caption{Average test F1-score $\pm$ standard deviation for the 1vs.1 datasets.}\label{ta:resultados2}}
\vspace*{2mm}
\resizebox{5.0in}{!}{
\centering
\begin{tabular}{llccc}
\hline
Normal class & \begin{tabular}{l} Anomalous class \end{tabular} & DAEF & OS-ELM & OC-SVM \\ 
\hline
Chocolate bars    & \begin{tabular}{c} Anise seeds \end{tabular} & 91.4$\pm$2.1 & 91.8$\pm$0.1  & 90.7$\pm$1.3  \\
Anise seeds    & \begin{tabular}{c} Chocolate bars \end{tabular} & 90.3$\pm$0.9 & 91.4$\pm$0.3  & 90.0$\pm$0.5  \\
Colored pencils    & \begin{tabular}{c} Ergonomic cushion \end{tabular} & 96.5$\pm$1.1 & 96.4$\pm$0.4  & 94.6$\pm$1.1  \\
Ergonomic cushion    & \begin{tabular}{c} Colored pencils \end{tabular} & 96.6$\pm$0.7 & 96.1$\pm$0.1  & 94.8$\pm$0.8  \\
Gaming mouse    & \begin{tabular}{c} PS4 membership \end{tabular} & 93.8$\pm$1.4 & 94.1$\pm$0.3  & 91.9$\pm$0.3  \\
PS4 membership  & \begin{tabular}{c} Gaming mouse \end{tabular} & 92.0$\pm$1.1 & 94.0$\pm$0.2  & 92.1$\pm$0.5  \\
Bluetooth speaker    & \begin{tabular}{c} Wi-Fi range extender \end{tabular} & 90.2$\pm$3.1 & 89.4$\pm$0.6  & 90.0$\pm$0.7  \\
Wi-Fi range extender    & \begin{tabular}{c}   Bluetooth speaker \end{tabular} & 93.0$\pm$0.5 & 92.9$\pm$0.2  & 92.3$\pm$0.4  \\
Foot insoles    & \begin{tabular}{c} Yoga leggings \end{tabular} & 90.4$\pm$1.0 & 92.2$\pm$0.2  & 90.5$\pm$0.4  \\
Yoga leggings    & \begin{tabular}{c} Foot insoles \end{tabular} & 89.9$\pm$2.2 & 90.9$\pm$0.3  & 90.4$\pm$0.6  \\
\hline
\end{tabular}
}
\end{table}

\subsection{Evaluating the explanations for model classifications} \label{evaluation2}

In Section \ref{method3}, we proposed as a method of explanation to support the model decisions one based on the occurrence of frequent terms, based on the hypothesis that normal reviews will tend to use certain terms regularly, while anomalous reviews will not. In this section, this approach is compared, qualitatively through user surveys, with two other popular alternative approaches to achieve such explainability, specifically SHAP \cite{NIPS2017_8a20a862} and GPT-3 \cite{https://doi.org/10.48550/arxiv.2005.14165}.

\subsubsection{Explanations based on SHAP} \label{results31}

Many explainability techniques base their operation on determining which characteristics of the dataset have most influenced the predictions. For example, in industrial scenarios, it is common for the features of the datasets to come directly from the physical aspects measured by the sensors of the machines, giving rise to variables such as temperatures, pressures or vibrations. By quantifying the influence of each of these variables on the output of the system we can achieve very useful explanations.

However, in the scenario proposed in this paper, the data received by the anomaly detection model as input are the reviews' embeddings, these being numerical vectors. The variables that make up the embeddings are not associated with aspects understandable by human beings such as temperatures or pressures, so determining which ones have influenced the most would not provide us with useful information.

To solve this problem, some explainability techniques have been adapted to deal with texts. In the case of NLP models, specifically transformers, SHAP (SHapley Additive exPlanations) \cite{NIPS2017_8a20a862} is one of the most widely used techniques. SHAP  is a game theory-based approach to explain the output of any machine learning model. SHAP also assigns each feature an importance value for a particular prediction, but it has been extended to provide interpretability to models that use embeddings as input. In these cases, it can quantify the importance of each word of the original text in the prediction, which generates a quite understandable interpretation for a human being. 

 \begin{figure}
\centerline{\includegraphics[width=\textwidth]{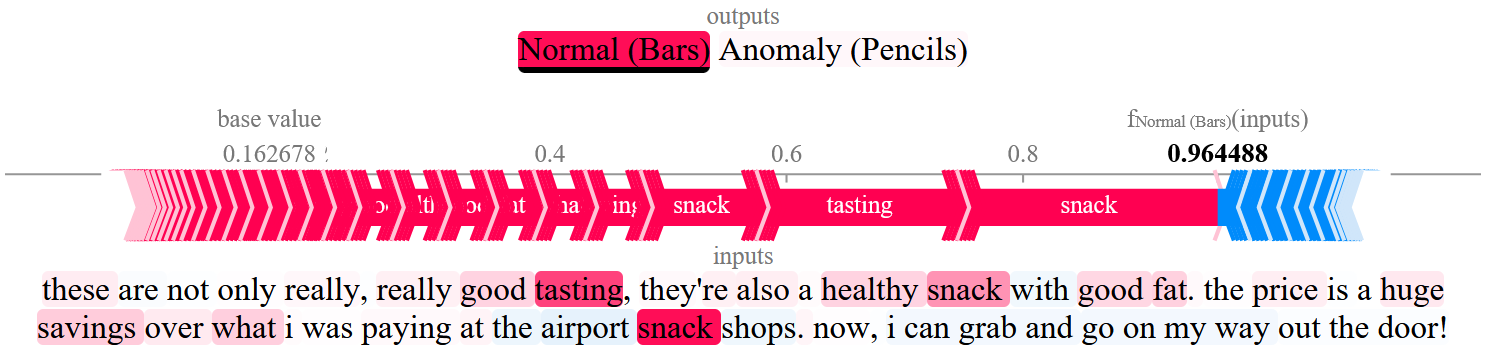}}
\caption{Explanation generated by SHAP for the anomalous reviews detection problem raised in this work. In this scenario, the reviews to be analysed correspond to the product "chocolate bars" (Bars). The review of this example was correctly classified as normal. The most influential terms in its classification as normal are marked in red, while the terms that promote the opposite class are highlighted in blue, in this case practically none since we are dealing with an obvious case. Greater intensity implies greater influence. In this case, the review terms that have most influenced its classification as normal are "snack" and "tasting".}
\label{shap}
\end{figure}

In this work, we considered that SHAP could be a good alternative to our proposal  to generate the explanations associated with the classification of the reviews. Figure \ref{shap} shows the explanation generated by SHAP for a review classified as normal, where it can be seen that each term of the review has an associated score representing its influence on the classification. Despite this, for the evaluation of explanation techniques and to facilitate the understanding of the explanation by the final users, during this work  (many of them unfamiliar with these techniques), the explanations generated by SHAP will be simplified, showing only the five most influential terms, instead of all.

\subsubsection{Explanations based on GPT-3} \label{results33}

There is no doubt that GPT-3 has started a technological revolution at all levels. Its availability to the general public has led to the discovery of a large number of unimaginable features before its release. The original idea was to create a high-level conversational bot, trained with a large amount of text available on the web, such as books, online encyclopedias or forums. However, its deep understanding of language has far exceeded the preset idea of a chatbot. GPT-3 is capable of successfully carrying out tasks that go beyond writing a joke or summarizing a novel, GPT-3 is capable of analyzing and developing code in multiple programming languages, generating SEO positioning strategies, or carrying out NLP tasks traditionally solved by ad hoc models, such as sentiment analysis.

\begin{figure}
\centerline{\includegraphics[width=\textwidth]{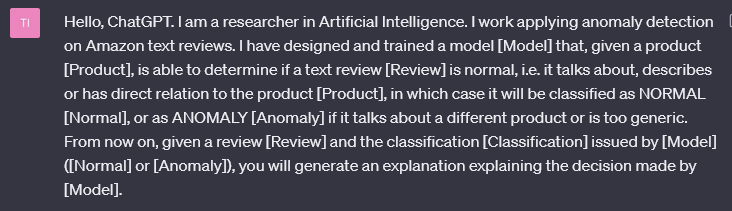}}
\caption{Initial prompt to present the anomaly detection problem to be solved to GPT-3.}
\label{GPT1}
\end{figure}

\begin{figure}
\centerline{\includegraphics[width=\textwidth]{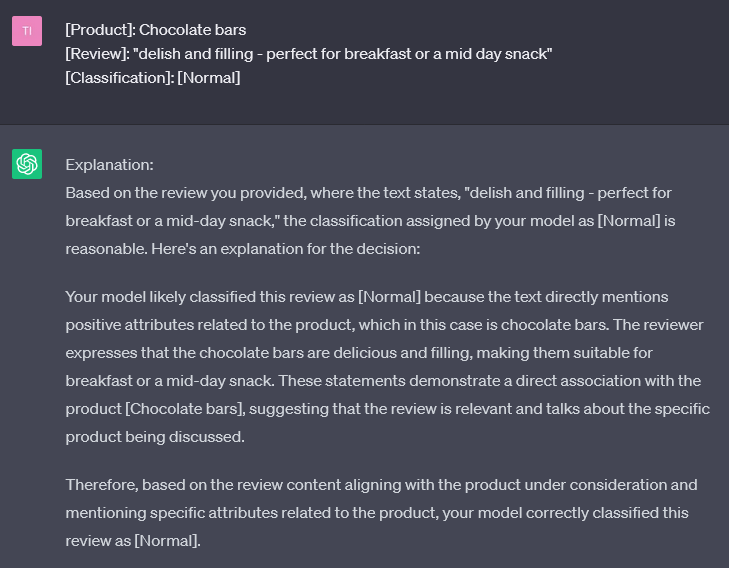}}
\caption{Prompt in which the classification of a review is explained by GPT-3. The product considered as normal is chocolate bars and the review has been classified as normal by the anomaly detection model.}
\label{GPT2}
\end{figure}

Due to its enormous potential, in this work, we have decided to study GPT-3 as an explainability model. To do this, first of all, we have introduced a prompt in which we describe the task that our anomaly detection model is carrying out, as well as the format with which we will work in future prompts (see Figure \ref{GPT1}). After this, and following the predefined format, GPT-3 is ready to generate the explanations (see Figure \ref{GPT2}). Although GPT-3 does not have direct knowledge of the anomaly detection model, its ability to generate consistent and intuitive responses can be of great help to humans reading the explanations.

We should note that since GPT-3 trains on a wide range of data from the internet, including data that may contain biases, there is a risk that the generated explanations may reflect or amplify those biases. It is essential to exercise caution and perform a critical analysis of the explanations generated, considering their context and possible inherent biases.

\subsubsection{How to evaluate explainability} \label{results4}

Unlike other tasks such as anomaly detection, the quality of the explanations generated to provide certain interpretability to a model is not easily measurable. In a scenario like the one described in this paper, the subjectivity of the users plays a strong role when determining whether an explanation is appropriate or not. Due to this, we have decided to carry out a comparative study of the three explainability techniques by means of a survey. This survey was disseminated through the students, professors, and research and administrative staff of our university, giving rise to a total of 241 participants. Table \ref{ta:AreasConocimientos} shows the number of participants by area of knowledge. To build the survey, reviews from the ``1 vs. 4'' scenario described in Table \ref{ta:resultados1} were used, considering ``chocolate bars'' as the normal class, and DAEF as an anomaly detection method.

\begin{table}[H]
\centering
{\caption{Area of knowledge of the survey participants.}\label{ta:AreasConocimientos}}
\vspace*{2mm}
\resizebox{3.0in}{!}{
\centering
\begin{tabular}{ll}
\hline
Knowledge area &  \begin{tabular}{l} Participants \end{tabular} \\
\hline
Engineering and Architecture &  \begin{tabular}{l} 87 (36.1\%) \end{tabular} \\ 
Social and Legal Sciences   & \begin{tabular}{l} 77 (32.0\%) \end{tabular} \\
Natural Sciences    & \begin{tabular}{l}  33 (13.7\%) \end{tabular} \\
Arts and Humanities    & \begin{tabular}{l} 23 (9.5\%) \end{tabular} \\
Health Sciences    & \begin{tabular}{l} 17 (7.1\%) \end{tabular} \\
Others    & \begin{tabular}{l} 4 (1.7\%) \end{tabular} \\
\hline
\end{tabular}
}
\end{table}

The survey consists of two different tests: (1) Forward simulation \cite{Hase2020EvaluatingEA}, which allows measuring the effect of explanations on users; (2) Personal utility, which allows measuring the utility of the explanations based on the personal preferences of the users.

\subsubsection{Forward simulation} \label{results41}

This test is inspired by the work carried out by P. Hase et al. \cite{Hase2020EvaluatingEA} and is divided into four phases: a Learning phase, a Pre-prediction phase, a Learning phase with explanations, and a Post-prediction phase. To begin, users are given 20 examples from the model's validation set with reviews and model predictions but no explanations. Then they must predict the model output for 10 new reviews. Users are not allowed to reference the learning data while in the prediction phases. Next, they return to the same learning examples, now with explanations included. Finally, they predict model behavior again on the same instances from the first prediction round. The classes of reviews chosen for each of the phases described are balanced between normal and abnormal. By design, any improvement in user performance in the Post prediction phase is attributable only to the addition of explanations that help the user to understand the model behaviour. 

\begin{figure}[H]
\centerline{\includegraphics[width=\textwidth]{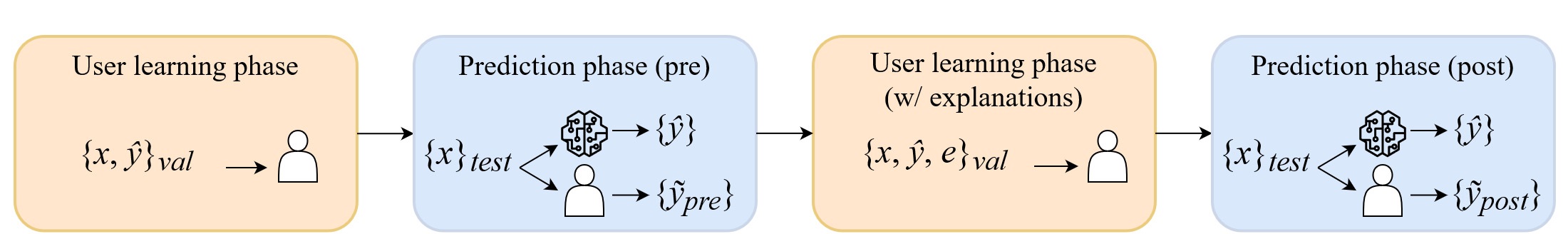}}
\caption{Forward simulation test procedure to measure human users’ ability to understand and predict
model behavior. To isolate the impact of explanations, baseline accuracy is measured first, followed by accuracy measurement after users have access to explanations of the model's behavior. The explained examples are different from the test instances.}
\label{explainability_forward}
\end{figure}

Figure \ref{explainability_forward} represents this procedure, where $x$ represents a review, $\widehat{y}$ is the class predicted by the model, and $\tilde{y}$ the class predicted by the human simulation. Taking this into account, the Explanation Effect can be calculated as follows:

\begin{equation}
Explanation\,  E\!f\!fect = P\! ost \; Accuracy - Pre \; Accuracy
\label{explanation}
\end{equation}

\noindent where the pre and post accuracies area calculated comparing the user's prediction against the model's prediction.
In order not to bias the results, we have decided that each person surveyed will only participate in the forward simulation of a single explainability technique, so that the total number of participants was divided randomly into three groups, each one associated with one of the three techniques: Terms frequency (78), SHAP (89), and GPT-3 (77). The three groups of participants will deal with the same reviews throughout the four phases of the test, only the explanations presented in the third phase of the test (learning with explanations) will vary depending on the assigned group/technique. Throughout the survey, we have warned users several times that they should try to simulate the behavior of the anomaly detection model, instead of ranking the reviews using their personal criteria.

Tables \ref{ta:test11} and \ref{ta:test12} show the results of this test. As can be seen in the first table, the average explanation effect of the three explainability techniques has not been very remarkable. The high standard deviation suggests a high variability between the different study participants, both in the initial (Pre) and subsequent (Post) classifications and therefore in the explanation effect. The initial difference between the groups in the Pre-accuracy and the closeness of the three techniques in the explanation effect does not allow us to opt for any of the three options.

In Table \ref{ta:test12} we can see the results of the test based on the area of knowledge of the participants. The results obtained using the different explainability techniques have been aggregated. The purpose of this comparison is to analyze whether there is any relationship between the technical background of the respondents and their performance on the test. As can be seen, there are slight differences between the values of pre and post accuracy, with the group of respondents belonging to the area of natural sciences standing out, whose explanation effect was the only positive one (4.5\%).

\begin{table}[H]
\centering
{\caption{Forward simulation tests.  Average accuracy and Explanation effect ($\pm$ standard deviation) for each explainability technique.}\label{ta:test11}}
\vspace*{2mm}
\resizebox{4.0in}{!}{
\centering
\begin{tabular}{lccc}
\hline
Technique &  \begin{tabular}{c} Pre \\ Accuracy \end{tabular} & \begin{tabular}{c} Post \\ Accuracy \end{tabular} & \begin{tabular}{c} Explanation \\ Effect \end{tabular} \\ 
\hline
Terms frequency    & \begin{tabular}{l} 76.2 $\pm$13.0 \end{tabular} & 72.8$\pm$13.8 & -3.4$\pm$12.7    \\
SHAP     & \begin{tabular}{l} 72.4$\pm$14.9 \end{tabular} & 71.9$\pm$14.9 & -0.5$\pm$13.8    \\
GPT-3   & \begin{tabular}{l} 69.7$\pm$15.2 \end{tabular} & 70.1 $\pm$17.1 & 0.4$\pm$14.3    \\
\hline
\end{tabular}
}
\end{table}

\begin{table}[H]
\centering
{\caption{Forward simulation tests. Average accuracy and Explanation effect ($\pm$ standard deviation) broken down by participants' area of knowledge.}\label{ta:test12}}
\vspace*{2mm}
\resizebox{4.8in}{!}{
\centering
\begin{tabular}{lccc}
\hline
Area of knowledge &  \begin{tabular}{c} Pre \\ Accuracy \end{tabular} & \begin{tabular}{c} Post \\ Accuracy \end{tabular} & \begin{tabular}{c} Explanation \\ Effect \end{tabular} \\ 
\hline
Engineering and Architecture    & \begin{tabular}{c} 76.3 $\pm$14.5 \end{tabular} & 73.9$\pm$16.1 & -2.4$\pm$12.9    \\
Social and Legal Sciences     & \begin{tabular}{c} 72.2$\pm$13.0 \end{tabular} & 69.5$\pm$15.8 & -2.7$\pm$13.0    \\
Natural Sciences   & \begin{tabular}{c} 70.9$\pm$14.7 \end{tabular} & 75.4 $\pm$12.8 & 4.5$\pm$13.5    \\
Arts and Humanities   & \begin{tabular}{c} 68.7$\pm$17.7 \end{tabular} & 67.8 $\pm$14.8 & -0.9$\pm$19.0    \\
Health Sciences   & \begin{tabular}{c} 72.0$\pm$14.2 \end{tabular} & 71.3 $\pm$10.6 & -0.7$\pm$10.3    \\
\hline
\end{tabular}
}
\end{table}

Analyzing the results we can affirm that the initial accuracy (pre-accuracy) is quite high, which indicates that the respondents tend to successfully reproduce the behavior of the model even if they do not have the explanations. This could be because, in this case, both the input data to the AD model (in natural language) and the problem it solves are easily understandable to a human, which means that the respondents are able to solve the classification problem by themselves. The standard deviation accompanying the pre-accuracy results is notable but does not become too high, which reaffirms the previous argument.

After supplying the respondents with the explanations associated with the reviews, the post-accuracy obtained by them presents values very similar to the previous scores (pre-accuracy). We can therefore affirm that in general terms the effect of the explanations, in this case, has not been beneficial for the users during this test.

The non-improvement may be due to several reasons. One of them may be the presence of reviews that show a certain degree of ambiguity, which not only makes their classification difficult for the respondents, but also for the AD models. However, in real scenarios the occurrence of reviews whose normality score is around the threshold value would be something to be expected, not all events are easily classifiable. Another possible reason may be that the tendency of some users throughout the survey has been to classify the reviews using their personal criteria, rather than trying to simulate the behavior of the anomaly detection model.

\subsubsection{Personal utility} \label{results42}

After completing the first test, the participants were given a second exercise, common to all participants, regardless of the group to which they were assigned in the previous phase. This consists of a subjective evaluation of the three explainability techniques. The idea is to present the participant with a review, its classification by the model, and an explanation generated by each of the three explainability techniques. The participant must order the explanations based on how useful it is to understand the reasoning behind the model's decision (ties were allowed between explanations). This process was repeated with a total of eight reviews. Tables \ref{ta:test21} and \ref{ta:test22} show the final average rankings of the explainability techniques.

\begin{table}[H]
\centering
{\caption{Personal utility tests. Average ranking ($\pm$ standard deviation) for each explainability technique.}\label{ta:test21}}
\vspace*{2mm}
\resizebox{2in}{!}{
\centering
\begin{tabular}{lc}
\hline
Technique &  \begin{tabular}{c} Position \end{tabular}  \\ 
\hline
Terms frequency    & \begin{tabular}{l} 1.6 $\pm$0.4 \end{tabular}  \\
SHAP     & \begin{tabular}{l} 2.1$\pm$0.5 \end{tabular}  \\
GPT-3   & \begin{tabular}{l} 1.7$\pm$0.5 \end{tabular} \\
\hline
\end{tabular}
}
\end{table}

As can be seen in Table \ref{ta:test21}, the explanations based on Term frequency (1.6) and GPT-3 (1.7) are in very close positions, both being above the third method SHAP (2.1). It is possible that the preference of respondents for the first two methods is due to the fact that the format of their explanations is more accessible and descriptive for a larger part of the population. Grouping the results by areas of knowledge (Table \ref{ta:test22}), we can see that the general trend is maintained for most areas. We can highlight the case of Social and Legal Sciences and Health Sciences, areas in which the positions in the ranking of Terms frequency and GPT-3 techniques are slightly inverted, with GPT-3 being the preferred option.

\begin{table}[H]
\centering
{\caption{Personal utility tests. Average ranking ($\pm$ standard deviation) for each technique and area of knowledge.}\label{ta:test22}}
\vspace*{2mm}
\resizebox{4in}{!}{
\centering
\begin{tabular}{llc}
\hline
Area of knowledge & Technique &  \begin{tabular}{c} Position \end{tabular}  \\ 
\hline
& Terms frequency    & \begin{tabular}{l} 1.6 $\pm$0.4 \end{tabular}  \\
Engineering and Architecture  & SHAP    & \begin{tabular}{l} 2.2 $\pm$0.5 \end{tabular}  \\
 & GPT-3    & \begin{tabular}{l} 1.7 $\pm$0.5 \end{tabular}  \\
\hline
 & Terms frequency     & \begin{tabular}{l} 1.7$\pm$0.4 \end{tabular}  \\
Social and Legal Sciences & SHAP     & \begin{tabular}{l} 1.9$\pm$0.4 \end{tabular}  \\
 & GPT-3    & \begin{tabular}{l} 1.6$\pm$0.5 \end{tabular}  \\
\hline
 & Terms frequency  & \begin{tabular}{l} 1.6$\pm$0.4 \end{tabular} \\
Natural Sciences & SHAP   & \begin{tabular}{l} 2.0$\pm$0.5 \end{tabular} \\
 & GPT-3   & \begin{tabular}{l} 1.7$\pm$0.5 \end{tabular} \\
\hline
 & Terms frequency   & \begin{tabular}{l} 1.5$\pm$0.4 \end{tabular} \\
Arts and Humanities & SHAP   & \begin{tabular}{l} 1.9$\pm$0.5 \end{tabular} \\
 & GPT-3   & \begin{tabular}{l} 1.8$\pm$0.5 \end{tabular} \\
\hline
  & Terms frequency  & \begin{tabular}{l} 1.8$\pm$0.3 \end{tabular} \\
Health Sciences  & SHAP   & \begin{tabular}{l} 2.1$\pm$0.5 \end{tabular} \\
 & GPT-3   & \begin{tabular}{l} 1.6$\pm$0.5 \end{tabular} \\

\hline
\end{tabular}
}
\end{table}

\section{Conclusion}

In this work, we have proposed a pipeline for detecting anomalous reviews associated with Amazon products, which can be directly extrapolated to other online review platforms or scenarios with similar characteristics. The representation of the reviews using MPNet embeddings has enabled the training of classical anomaly detection algorithms that have achieved a very good performance. These have been evaluated using reviews from different products and categories, and the score they emit allows us to sort the reviews based on their normality. 

A technique based on the occurrence of frequent terms has been proposed to generate explanations associated with the classifications of the reviews. This technique has been compared with SHAP, one of the reference post-hoc techniques in the field of explainability, and with GPT-3, due to its high power and versatility. To evaluate this aspect of the pipeline, we conducted a two-part survey in which 241 members of the university community participated. 


From the first part of the explainability test we can conclude that, in general terms, the effect of the explanations has not been beneficial for the users. In any case, these tests allow us to reflect on the difficulty of using explainability and evaluation techniques in borderline scenarios where subjectivity plays an important role, such as the one presented in this article or in other fields of NLP, as well as in areas such as image or audio generation.

Regarding the second part of the explainability test, we have been able to conclude that respondents preferred explanations that presented a more natural and familiar appearance over more condensed and concise explanations such as those provided by SHAP, regardless of the explanation effect they provide. Explanations based on term frequency analysis have been preferred by respondents along with GPT-3, however, our approach presents a significantly lower computational costs and both its use and the explanations produced are simpler for the users.

As future work, it would be interesting to evaluate GPT-3 or other large models carrying out the complete process followed by the pipeline proposed in this work, instead of being tested only in the explainability module. We have not carried out this test due to the high computational cost that would be involved in processing the thousands of reviews to be evaluated using GPT-3. Another interesting possible line of future work would be to broaden the scope of the survey, both in terms of the number of products involved and the number of reviews, in order to clarify the conclusions reached at the forward simulation stage. Lastly, it would be very useful to try presenting the explanations issued by SHAP in a more familiar or natural format for the end user, so that we can see if their level of preference is increased for the general public.

\section*{Acknowledgements}

This work was supported in part by grant \textit{Machine Learning on the Edge - Ayudas Fundaci\'on BBVA a Equipos de Investigaci\'on Cient\'ifica 2019}; the Spanish National Plan for Scientific and Technical Research and Innovation (PID2019-109238GB-C22 and TED2021-130599A-I00); the Xunta de Galicia (ED431C 2022/44) and ERDF funds. CITIC, as a Research Center of the University System of Galicia, is funded by Consellería de Educación, Universidade e Formación Profesional of the Xunta de Galicia, Spain through the European Regional Development Fund (ERDF) and the Secretaría Xeral de Universidades (Ref. ED431G 2019/01).

\bibliographystyle{elsarticle-num} 
\bibliography{cas-refs}

\appendix
\section{Hyperparameters used during training} \label{appendix1}

This appendix contains the values of the hyperparameters finally chosen as the best for each method and dataset, listed in Tables \ref{ta:dinos2} and \ref{ta:dinos3}.

DAEF \cite{NOVOAPARADELA2023109805}, OS-ELM \cite{4012031}, and OC-SVM \cite{1437839} respectively.

\begin{itemize}
\item Deep Autoencoder for Federated learning (DAEF)\cite{NOVOAPARADELA2023109805}.
    \begin{itemize}
        \item Architecture: Neurons per layer.
        \item $\lambda_{hid}$: Regularization hyperparameter of the hidden layer.
        \item $\lambda_{last}$: Regularization hyperparameter of the last layer.
        \item $\mu$: Anomaly threshold.
    \end {itemize}

\item Online Sequential Extreme Learning Machine (OS-ELM)\cite{4012031}.
    \begin{itemize}
        \item Architecture: Neurons per layer.
        \item $\mu$: Anomaly threshold.
    \end {itemize}
    
\item One-Class Support Vector Machine (OC-SVM)\cite{1437839}.
    \begin{itemize}
        \item An upper bound on the fraction of training errors and a lower bound of the fraction of support vectors ($\nu$).
        \item Kernel type: Linear, Polynomial or RBF.
        \item Kernel coefficient $\gamma$ (in the case of polynomial and RBF kernels).
        \item Degree (in the case of polynomial kernel). 
    \end {itemize}
\end{itemize}

\begin{table}[H]
\centering
\begin{turn}{0}
\resizebox{5.0in}{!}{
\centering
\begin{tabular}{|c|c|c|c|}
\hline
Normal class & DAEF & OS-ELM & OC-SVM \\ 
\hline
Chocolate bars    & \begin{tabular}{c} Arch: [768, 550, 650, 768], \\ $\lambda_{hid}$: 0.9, $\lambda_{last}$: 0.9, \\ $\mu$: outlier IQR
  \end{tabular} & \begin{tabular}{c} Arch: [768, 400, 768], \\  $\mu$: extreme IQR
  \end{tabular} & \begin{tabular}{c} $\nu$: 0.1, \\ kernel: rbf, \\ $\gamma$: 0.00097
  \end{tabular} \\
\hline
Colored pencils  & \begin{tabular}{c} Arch:  [768, 550, 650, 768], \\ $\lambda_{hid}$: 0.1, $\lambda_{last}$: 0.1, \\ $\mu$: outlier IQR
  \end{tabular} & \begin{tabular}{c} Arch:  [768, 500, 768] ,  \\ $\mu$: extreme IQR
  \end{tabular} & \begin{tabular}{c} $\nu$: 0.1, kernel: poly, \\ degree: 2, \\ $\gamma$: scale
  \end{tabular} \\
\hline
Gaming mouse   & \begin{tabular}{c} Arch: [768, 550, 650, 768], \\ $\lambda_{hid}$: 0.1, $\lambda_{last}$: 0.1, \\ $\mu$: outlier IQR
  \end{tabular} & \begin{tabular}{c} Arch: [768, 100, 768],  \\$\mu$: outlier IQR
  \end{tabular} & \begin{tabular}{c} $\nu$: 0.1, \\ kernel: rbf, \\ $\gamma$: scale
  \end{tabular} \\
\hline
Bluetooth speaker  & \begin{tabular}{c} Arch: [768, 550, 650, 768], \\ $\lambda_{hid}$: 0.75, $\lambda_{last}$: 0.1, \\ $\mu$: outlier IQR
  \end{tabular} & \begin{tabular}{c} Arch: [768, 20, 768],  \\ $\mu$: outlier IQR
  \end{tabular} & \begin{tabular}{c} $\nu$: 0.1, \\ kernel: sigmoid, \\ $\gamma$: scale
  \end{tabular} \\
\hline
Foot insoles  & \begin{tabular}{c} Arch: [768, 550, 650, 768], \\ $\lambda_{hid}$: 0.9, $\lambda_{last}$: 0.9, \\ $\mu$: outlier IQR
  \end{tabular} & \begin{tabular}{c} Arch: [768, 300, 768]], \\ $\mu$: outlier IQR
  \end{tabular} & \begin{tabular}{c} $\nu$: 0.1, \\ kernel: rbf, \\ $\gamma$: scale
  \end{tabular} \\
\hline
\end{tabular}
}
\end{turn}
\caption{Hyperparameters used during the 1vs.4 experimentation.}
\label{ta:dinos2}
\end{table}

\begin{table}[H]
\centering
\begin{turn}{0}
\resizebox{5.0in}{!}{
\centering
\begin{tabular}{|c|c|c|c|}
\hline
Normal class & DAEF & OS-ELM & OC-SVM \\ 
\hline
Chocolate bars    & \begin{tabular}{c} Arch: [768, 550, 650, 768], \\ $\lambda_{hid}$: 0.1, $\lambda_{last}$: 0.1, \\ $\mu$: $Q_{90}$
  \end{tabular} & \begin{tabular}{c} Arch: [768, 400, 768], \\  $\mu$: outlier IQR
  \end{tabular} & \begin{tabular}{c} $\nu$: 0.1, kernel: poly, \\ degree: 3,  $\gamma$: scale
  \end{tabular} \\
\hline
Anise seeds  & \begin{tabular}{c} Arch:  [768, 550, 650, 768], \\ $\lambda_{hid}$: 0.9, $\lambda_{last}$: 0.9, \\ $\mu$: $Q_{80}$
  \end{tabular} & \begin{tabular}{c} Arch:  [768, 500, 768] ,  \\ $\mu$: outlier IQR
  \end{tabular} & \begin{tabular}{c} $\nu$: 0.1, kernel: poly, \\ degree: 4,  $\gamma$: scale
  \end{tabular} \\
\hline
Colored pencils   & \begin{tabular}{c} Arch: [768, 550, 650, 768], \\ $\lambda_{hid}$: 0.25, $\lambda_{last}$: 0.25, \\ $\mu$: outlier IQR
  \end{tabular} & \begin{tabular}{c} Arch: [768, 500, 768],  \\$\mu$: extreme IQR
  \end{tabular} & \begin{tabular}{c} $\nu$: 0.1, \\ kernel: rbf,  $\gamma$: scale
  \end{tabular} \\
\hline
Ergonomic cushion  & \begin{tabular}{c} Arch: [768, 550, 650, 768], \\ $\lambda_{hid}$: 0.5, $\lambda_{last}$: 0.1, \\ $\mu$: outlier IQR
  \end{tabular} & \begin{tabular}{c} Arch: [768, 500, 768],  \\ $\mu$: extreme IQR
  \end{tabular} & \begin{tabular}{c} $\nu$: 0.1, \\ kernel: sigmoid,  $\gamma$: scale
  \end{tabular} \\
\hline
Gaming mouse  & \begin{tabular}{c} Arch: [768, 550, 650, 768], \\ $\lambda_{hid}$: 0.9, $\lambda_{last}$: 0.9, \\ $\mu$: $Q_{90}$
  \end{tabular} & \begin{tabular}{c} Arch: [768, 500, 768]], \\ $\mu$: extreme IQR
  \end{tabular} & \begin{tabular}{c} $\nu$: 0.1, kernel: poly, \\ degree: 3,  $\gamma$: scale
  \end{tabular} \\
\hline
PS4 membership    & \begin{tabular}{c} Arch: [768, 550, 650, 768], \\ $\lambda_{hid}$: 0.1, $\lambda_{last}$: 0.1, \\ $\mu$: $Q_{90}$
  \end{tabular} & \begin{tabular}{c} Arch: [768, 400, 768], \\  $\mu$: extreme IQR
  \end{tabular} & \begin{tabular}{c} $\nu$: 0.1, \\ kernel: rbf,  $\gamma$: 0.00097
  \end{tabular} \\
\hline
Bluetooth speaker  & \begin{tabular}{c} Arch:  [768, 550, 650, 768], \\ $\lambda_{hid}$: 0.9, $\lambda_{last}$: 0.9, \\ $\mu$: $Q_{90}$
  \end{tabular} & \begin{tabular}{c} Arch:  [768, 500, 768] ,  \\ $\mu$: $Q_{90}$
  \end{tabular} & \begin{tabular}{c} $\nu$: 0.2, \\ kernel: linear,  $\gamma$: scale
  \end{tabular} \\
\hline
Wi-Fi range extender   & \begin{tabular}{c} Arch: [768, 550, 650, 768], \\ $\lambda_{hid}$: 0.25, $\lambda_{last}$: 0.1, \\ $\mu$: $Q_{90}$
  \end{tabular} & \begin{tabular}{c} Arch: [768, 500, 768],  \\$\mu$: outlier IQR
  \end{tabular} & \begin{tabular}{c} $\nu$: 0.1, kernel: poly, \\ degree: 4,  $\gamma$: scale
  \end{tabular} \\
\hline
Foot insoles  & \begin{tabular}{c} Arch: [768, 550, 650, 768], \\ $\lambda_{hid}$: 0.75, $\lambda_{last}$: 0.1, \\ $\mu$: outlier IQR
  \end{tabular} & \begin{tabular}{c} Arch: [768, 200, 768],  \\ $\mu$: outlier IQR
  \end{tabular} & \begin{tabular}{c} $\nu$: 0.1, kernel: poly, \\ degree: 4,  $\gamma$: scale
  \end{tabular} \\
\hline
Yoga leggings  & \begin{tabular}{c} Arch: [768, 550, 650, 768], \\ $\lambda_{hid}$: 0.9, $\lambda_{last}$: 0.9, \\ $\mu$: $Q_{90}$
  \end{tabular} & \begin{tabular}{c} Arch: [768, 500, 768]], \\ $\mu$: extreme IQR
  \end{tabular} & \begin{tabular}{c} $\nu$: 0.1, kernel: poly, \\ degree: 4,  $\gamma$: scale
  \end{tabular} \\
\hline
\end{tabular}
}
\end{turn}
\caption{Hyperparameters used during the 1vs.1 experimentation.}
\label{ta:dinos3}
\end{table}

\end{document}